\documentclass{article}

\usepackage[final,nonatbib]{tackling_climate_workshop_style}

\usepackage[utf8]{inputenc} 
\usepackage[T1]{fontenc}    
\usepackage{hyperref}       
\usepackage{url}            
\usepackage{booktabs}       
\usepackage{amsfonts}       
\usepackage{amsmath}        
\usepackage{nicefrac}       
\usepackage{microtype}      
\usepackage{xcolor}         
\usepackage{mathtools}
\usepackage{float}
\usepackage{rotating}       
\usepackage{array,multirow} 

\usepackage{caption}
\usepackage{subcaption}

\usepackage{tikz}
\usetikzlibrary{positioning, arrows}

\usepackage[ruled,vlined]{algorithm2e}
\usepackage{bm}

\usepackage[backend=bibtex,maxbibnames=30,sorting=none,date=year,doi=false,isbn=false,url=false]{biblatex}
\DeclareMathOperator*{\argmax}{arg\,max}

\DeclarePairedDelimiterX{\infdivx}[2]{(}{)}{%
	#1\;\delimsize\|\;#2%
}

\title{Identifying the atmospheric drivers of drought and heat using a smoothed deep learning approach}

\author{
  Magdalena Mittermeier\\
  Department of Geography\\
  LMU Munich\\
  \texttt{m.mittermeier@lmu.de} \\
  \And
  Maximilian Weigert\\
  Statistical Consulting StaBLab\\
  Department of Statistics\\
  LMU Munich\\
  \texttt{maximilian.weigert@stat.uni-muenchen.de} \\
  \And
  David R\"ugamer\\
  Department of Statistics\\
  LMU Munich\\
  \texttt{david.ruegamer@stat.uni-muenchen.de} \\
}

\begin{document}

\maketitle


\begin{abstract}
Europe was hit by several, disastrous heat and drought events in recent summers. Besides thermodynamic influences, such hot and dry extremes are driven by certain atmospheric situations including anticyclonic conditions. Effects of climate change on atmospheric circulations are complex and many open research questions remain in this context, e.g., on future trends of anticyclonic conditions. Based on the combination of a catalog of labeled circulation patterns and spatial atmospheric variables, we propose a smoothed convolutional neural network classifier for six types of anticyclonic circulations that are associated with drought and heat. Our work can help to identify important drivers of hot and dry extremes in climate simulations, which allows to unveil the impact of climate change on these drivers. We address various challenges inherent to circulation pattern classification that are also present in other climate patterns, e.g., subjective labels and unambiguous transition periods.
\end{abstract}



\section{Introduction}

%
%
%

In recent summers such as those of 2003, 2010 and 2018, Europe has been subject to particularly outstanding summer drought and heat events, which caused large economic and societal damage including heat-related deaths \cite{Bastos2020, Liu2020}.
The frequency and intensity of hot and dry extremes has recently increased and is projected to further increase due to climate change and rising global mean temperatures \cite{Suarez-Gutierrez2020, Spinoni2020}.

\paragraph{Drivers of hot and dry extremes} There are two key processes leading to drought and heat events: thermodynamic and dynamic factors. Thermodynamic factors involve, e.g., evaporation and the feedback between soil moisture and air temperature. Dynamic factors on the other hand describe the atmospheric drivers of heat and drought, which are mainly anticyclonic conditions and blocking \cite{Suarez-Gutierrez2020}. While anticyclonic conditions go along with various high-pressure systems, blocking describes a particular, persistent high-pressure situation that is associated with the displacement of westerly winds and their accompanying weather systems \cite{Suarez-Gutierrez2020, IPCC2021}. These atmospheric drivers of hot and dry extremes are part of the large-scale atmospheric circulation in the mid-latitudes, which control the weather and climate over Europe \cite{Huguenin2020, Woollings2010}. Changes in the atmospheric circulation are complex with opposing processes and thus many open research questions remain \cite{IPCC2021, Stendel2021}. 

\paragraph{Classification of circulation patterns} Objectively classifying the circulation patterns that are associated with hot and dry extremes is an important step towards a better understanding of how climate change affects the atmospheric drivers of potentially disastrous extreme events. Previous studies, e.g., \cite{Mittermeier2019, Racah2016, Kurth2017} have shown that deep learning approaches are powerful tools for the detection of extreme weather in climate simulations. In this study, we use a subjective catalog of circulation type classifications over Europe by Hess \& Brezowsky \cite{Hess1969, Werner2010}. Our goal is to learn the categorization of six circulation types with anticyclonic conditions over Europe, which are associated with dry and hot summer conditions in Central Europe \cite{KLIWA2012}.
The classification of circulation types comes with various challenges that need to be addressed with appropriate modeling strategies. Challenges include noisy labels due to subjective expert choices in ambiguous climate situations \cite{Huth2008}, an imbalanced class distribution of labels, undefined transition days between successive circulation patterns, and a fixed dwell time of a circulation pattern of at least three consecutive days by its definition \cite{Werner2010}. 

\paragraph{Our contribution} In this work we propose a novel modeling procedure to address existing challenges in classifying anticyclonic circulation patterns. Especially in times of large ensembles of climate simulations \cite{Maher2021} that consist of dozens of model runs and thousands of model years, our study can help to efficiently analyse large climate simulations and be another piece of the puzzle to better understand changes in the atmospheric drivers of drought and heat. 


\section{Data}

\begin{figure}[htbp]
    \includegraphics[width=\textwidth, trim=0.4cm 0 0 0]{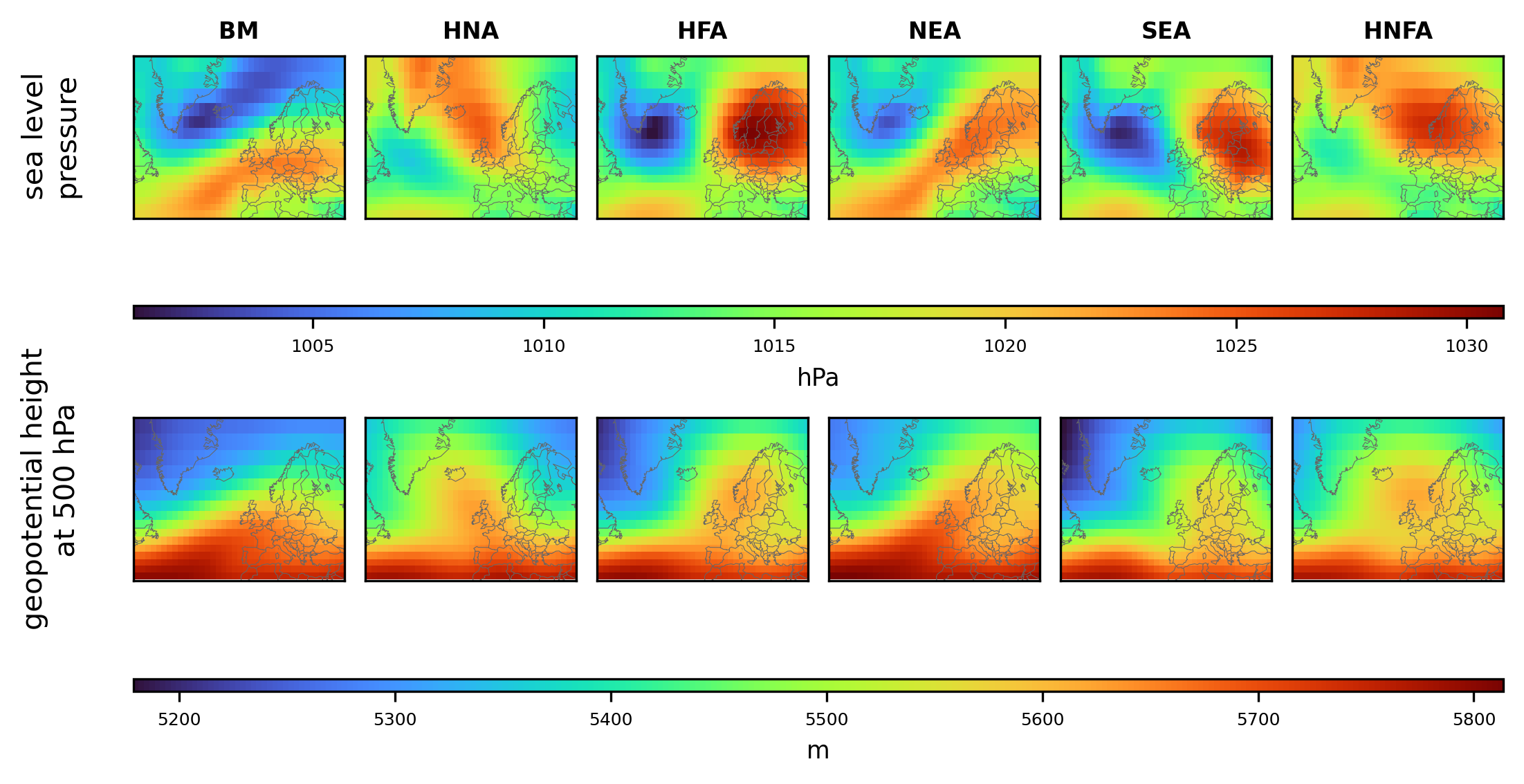}
    \caption{Mean air pressure patterns of the six anticyclonic circulation types BM, HNA, NEA, HFA, HNFA, and SEA (columns) averaged over all days in the period between 1900 and 2010. The plots are shown for the variables (rows) sea level pressure [hPa] and geopotential height at 500 hPa [m].
    }
    \label{fig:plots-gwl}
\end{figure}

The Hess \& Brezowsky catalog contains a subjective categorization of circulation patterns created by experts manually labelling air pressure patterns over Europe into 29 classes. In this way, daily air pressure constellations are retrospectively assigned to one of these classes. A circulation pattern is, by definition, required to last at least three days. The six circulation patterns associated with heat and drought are (abbreviations originate from German): \textit{Zonal ridge across Central Europe (BM), Norwegian Sea-Iceland high, anticyclonic (HNA), North-easterly anticyclonic (NEA), Fennoscandian high, anticyclonic (HFA), Norwegian Sea-Fennoscandian high, anticyclonic (HNFA), and South-easterly anticyclonic (SEA)} \cite{sykorova2020, KLIWA2012}. The mean air pressure patterns for the six circulation patterns of interest are given in Figure~\ref{fig:plots-gwl} for the variables \emph{sea level pressure} and \emph{geopotential height at 500 hPa} (average values at roughly 5500 meters height). For the analysis of heat and drought, the remaining 23 circulation types are assigned to a residual class. The frequencies of the six anticyclonic patterns are between 8.5\% (BM) and 1.4\% (HNFA), whereas the residual class comprises about 80\% of the days.


Next to the catalog by Hess \& Brezowsky, we supplement our data base with the ERA-20C reanalysis data by the European Centre for Medium-Range Weather Forecasts \cite{poli2016}. The long record ERA-20C data set contains global spatial information on various climate parameters from 1900 to 2010. In accordance with the definition by \cite{Werner2010}, we use the two atmospheric variables, sea level pressure and geopotential height at 500 hPa as predictors for circulation patterns. The spatial domain of our data set is defined over a region covering Europe and the North Atlantic \cite{Werner2010} (see Figure \ref{fig:plots-gwl}) with a spatial resolution of 5\textdegree x 5\textdegree~resulting in a grid of 16 x 29 pixels. The resulting data set contains daily information over 111 years, i.e., T = 40541 observations.

\section{Methods}

\paragraph{Model definition, training, tuning and evaluation} To account for the spatial information and the specific characteristics of circulation patterns, we use a convolutional neural network (CNN) following \cite{liu2016}, who propose a network architecture for climate pattern detection problems. Since air pressure patterns of atmospheric features are comparatively simple,
our chosen architecture 
consists of only two convolutional layers with larger kernels (5x5-8 and 5x5-16), a dropout layer and two fully-connected layers as well as individual channels for both climate parameters. The two different atmospheric variables are included as individual channels in the CNN as in \cite{Liu2020, Racah2016}. While there is reason to believe that accounting for the temporal structure of our data, e.g., through a structured model \cite{Ruegamer2021} or ConvLSTM \cite{Shi.2015} improves the model, a previous study \cite{henri2021new} showed no improvement in the classification of circulation patterns when using a temporal-aware architecture. As explained in the next paragraph, our approach instead smoothes predicted labels to account for their temporal nature.

The model is trained using Adam optimization with a batch size of 128 for 35 epochs and early stopping based on a validation set of size 3650 with patience of 6 epochs. Hyperparameter tuning for learning rate and dropout rate is performed using Bayesian optimization \cite{snoek2012}. We evaluate the model using overall accuracy and macro F1-score. For class-specific evaluations, we consider recall and precision. To obtain performance estimates that are as unbiased as possible, a nested cross-validation with ten inner and eleven outer folds is used. In order to not leak intra-year information, observations within the same year are required to belong to the same fold.

\paragraph{Modeling challenges} Our approach takes into account several data-specific characteristics for circulation pattern data. First, we employ a loss-weighting scheme to account for imbalanced classes by weighting the classes with their inverse frequencies. Moreover, the assigned categories in the Hess \& Brezowsky catalog can be noisy, in particular for transition days between two subsequent circulation patterns. This is due to the continuous movement of pressure systems while circulation types are discrete by definition and in-between states do not fit in one or the other class. We address this problem by using label-smoothing \cite{szegedy2016} for the first and last day of each occurrence of a specific circulation pattern. Finally, our target variable must adhere to the aforementioned definition of a circulation pattern, implying a pattern to last at least three days. A transition-smoothing step ensures that this three day rule is respected. In this step, the final predicted class $\Tilde{y}_t$ at time point $t = 2, ..., T - 2$, is given by
\begin{equation*}
\tilde{y}_t =
\begin{cases}
\hat{y}_{t-1} & \text{if } \hat{y}_{t-1} = \hat{y}_{t+1} \text{ (Neighborhood Consistency)}, \\
\hat{y}_{t-1} & \text{if } \hat{y}_t = \hat{y}_{t+1} \land \hat{y}_{t-1} = \hat{y}_{t+2} \text{ (2-days Consistency)}, \\
m(\hat{\boldsymbol{\pi}}_{t-1}, \hat{\boldsymbol{\pi}}_{t+1}) & \text{if } \hat{y}_t \neq \hat{y}_{t+1} \land \hat{y}_{t-1} \neq \hat{y}_{t+1} \text{  (Transition Membership)}, \\
m(\hat{\boldsymbol{\pi}}_{t-1}, \hat{\boldsymbol{\pi}}_{t+2}) & \text{else}, \\
\end{cases}
\end{equation*}
where $\hat{\boldsymbol{\pi}}_t$ denotes the predicted probability vector at time $t$, $\hat{y}_{t} = \argmax \hat{\boldsymbol{\pi}}_t$ the predicted class prior to the transition-smoothing step, and 
\begin{equation*}
m(\boldsymbol{\pi}_s, \boldsymbol{\pi}_t) = \argmax \{ \boldsymbol{\pi}_{u^\ast} \} \text{ with } u^\ast = \argmax_{u \in \{s,t\}} \{ \max(\boldsymbol{\pi}_u) \}   .
\end{equation*} 
This guarantees consistency with the required three day rule and systematically replaces isolated single or two day-type predictions.


\section{Results}

Taking into account the aforementioned subjectivity of the circulation pattern catalog and the noisy labels, the overall performance of the proposed model is satisfactory and our proposed smoothing approaches consistently improve the model across all classes.
Our smoothed convolutional neural network classifier achieves an macro F1-score of 38.4\% and an overall accuracy of 59.9\% averaged over the test sets during nested cross-validation.  Table~\ref{tab:confusion_matrix} shows the corresponding confusion matrix together with the precision and recall. 
The best performance in terms of recall is achieved for the circulation patterns HNA and BM, the lowest performance for NEA. In absolute numbers, misclassifications mainly occur for residual class observations. Due to the proposed error weighting technique, we obtain larger recall than precision values except for the residual class.

\begin{table}[h]
\small
\begin{center}
\caption[confusion matrix]{Confusion matrix of our proposed smoothed approach, averaged over the test sets in the nested cross-validation. Correctly classified classes are highlighted in bold.}
\label{tab:confusion_matrix}
\begin{tabular}{p{0.1cm} l c c c c c c c | c c c c}
\toprule
\\[-2ex]
&&\multicolumn{7}{c}{LABELS}& \\[0.5ex]
\midrule
\\[-2ex]
 & & BM & HNA & HFA & NEA & SEA & HNFA & Residual & $\sum$ & Precision \\[0.5ex]
\cline{2-11}
\parbox[b]{1mm}{\multirow{11}{*}{\rotatebox[origin=b]{90}{ OUTPUTS}}} &&&& \\[-2ex] 
\\[-2ex]
 & BM & \textbf{208.8} & 4.0 & 7.7 & 6.5 & 1.7 & 5.0 & 477.4 & 711.2 & 0.29 \\[1ex]
 & HNA & 11.9 & \textbf{75.8} & 3.7 & 4.5 & 7.1 & 3.7 & 204.9 & 311.5 & 0.24 \\[1ex]
 & HFA & 22.9 & 3.5 & \textbf{41.9} & 14.6 & 3.6 & 1.5 & 138.6 & 226.6 & 0.18 \\[1ex]
 & NEA & 10.4 & 2.8 & 10.4 & \textbf{61.5} & 6.3 & 11.7 & 85.9 & 188.9 & 0.33 \\[1ex]
 & SEA & 3.2 & 11.6 & 3.6 & 15.0 & \textbf{25.2} & 5.5 & 78.4 & 142.5 & 0.18 \\[1ex]
 & HNFA & 9.1 & 3.8 & 2.4 & 20.5 & 5.5 & \textbf{44.7} & 185.7 & 271.7 & 0.16 \\[1ex]
 & Residual & 43.3 & 7.0 & 6.6 & 5.2 & 2.7 & 5.3 & \textbf{1729.9} & 1800 & 0.96 \\[1ex]\hline
& & & & & & & & & & \\ 
& $\sum$ & 309.6 & 108.6 & 76.2 & 127.7 & 52.1 & 77.4 & 2900.7 & \textbf{3652.4} & -- \\[1ex]
& Recall & 0.67 & 0.70 & 0.55 & 0.48 & 0.48 & 0.58 & 0.60 & --  & -- \\[0.5ex]
\bottomrule
\end{tabular}
\end{center}
\end{table}

Table~\ref{tab:comparison} shows an ablation study of our modeling procedure. Including label- and transition-smoothing 
improves the overall accuracy by 4 percentage points and the macro F1-score by 2 percentage points compared to a model without any smoothing steps. 
The class specific F1-scores also considerably increase for all patterns. A comparison of the networks without label-smoothing and without transition-smoothing indicates that the label-smoothing step has rather little impact while the proposed transition-smoothing is the key to our observed performance gains.

\begin{table}[h]
\small
\begin{center}
\caption[comparison measures]{Comparison of class-specific F1-scores (first 7 columns), accuracy and macro F1-Score (last two columns) for the final smoothed model (Final), a model without label-smoothing (NO LS), a model without transition-smoothing (No TS) and a model without label-smoothing (No LS and TS). Best results are highlighted in bold.}
\label{tab:comparison}
\begin{tabular}{p{0.1cm} l c c c c c c c | c c c}
\toprule
\\[-2ex]
 & & BM & HNA & HFA & NEA & SEA & HNFA & Residual & \textbf{Accuracy} & \textbf{F1-score} \\[0.5ex] 
\cline{2-11}
\parbox[b]{1mm}{\multirow{6}{*}{\rotatebox[origin=b]{90}{MODEL}}} &&&& \\[-2ex] 
\\[-2ex]
& Smoothed network & \textbf{0.41} & \textbf{0.36} & \textbf{0.28} & {0.39} & \textbf{0.26} & \textbf{0.26} & \textbf{0.74} & \textbf{0.60} & {0.38} \\[1ex]
& No LS & \textbf{0.41} & \textbf{0.36} & \textbf{0.28} & \textbf{0.40} & \textbf{0.26} & \textbf{0.26} & \textbf{0.74} & \textbf{0.60} & \textbf{0.39} \\[1ex]
& No TS & {0.39} & {0.33} & {0.25} & {0.36} & {0.23} & {0.23} & {0.70} & {0.56} & {0.36} \\[1ex]
& No LS and TS & 0.39 & 0.33 & 0.25 & 0.37 & 0.23 & 0.23 & 0.70 &
{0.56} & {0.36} \\[1ex] 
\bottomrule
\end{tabular}
\end{center}
\end{table}

\section{Conclusion and Outlook}
Our results indicate the high potential of deep learning-based methods in classifying the atmospheric drivers of drought and heat. We also demonstrate the effectiveness of our smoothed approach to deal with typical challenges in circulation type classifications, e.g., transition-smoothing for historical dwell time definitions. To the best of our knowledge, we are the first to use air pressure patterns over Europe to classify circulation patterns associated with drought and heat as given in the Hess \& Brezowsky catalog. While the proposed approach can potentially also be used for other circulation patterns associated with different kinds of extreme climate events, our goal was to establish a baseline model for this specific and highly relevant circulation pattern categorization. Although our approach relies on a network architecture developed for climate applications \cite{liu2016}, there is room for improvement in modeling the analyzed patterns. 
As an alternative to the transition-smoothing step, we will investigate a deep hidden Markov model that accounts for the state dwell times by assuming a latent process that emulates the data-inherent three day transition rule.


\section*{Acknowledgements}
We thank the anonymous reviewers for their constructive comments which helped us to improve the manuscript.
The work of MM is funded through the ClimEx project (www.climex-project.org) by the Bavarian State Ministry for the Environment and Consumer Protection, the work of MW and DR by the German Federal Ministry of Education and Research (BMBF) under Grant No. 01IS18036A. 
The provision of the Hess \& Brezowsky catalog for the years 1900-2010 by the German Weather Service is highly appreciated.
The authors of this work take full responsibility of its content. 

\printbibliography
\end{document}